\documentclass[10pt,twocolumn,letterpaper]{article}

\usepackage{iccv}
\usepackage{times}
\usepackage{epsfig}
\usepackage{graphicx}
\usepackage{amsmath}
\usepackage{amssymb}
\usepackage{booktabs}
\usepackage{enumitem}
\usepackage{wrapfig}

\usepackage[breaklinks=true,bookmarks=false]{hyperref}

\iccvfinalcopy 

\def\iccvPaperID{***} 

\ificcvfinal\fi

\begin{document}

\title{NeRF-GAN Distillation for Efficient 3D-Aware Generation with Convolutions}

\author{
Mohamad Shahbazi\textsuperscript{\rm 1} \and Evangelos Ntavelis\textsuperscript{\rm 1,3} \and Alessio Tonioni\textsuperscript{\rm 2} \and Edo Collins\textsuperscript{\rm 2} \and Danda Pani Paudel\textsuperscript{\rm 1} \and Martin Danelljan\textsuperscript{\rm 1} \and  Luc Van Gool\textsuperscript{\rm 1} \and \\
\textsuperscript{\rm 1}Computer Vision Lab, ETH Z\"urich\quad\textsuperscript{\rm 2}Google Z\"urich\quad\textsuperscript{\rm 3}ML \& Robotics, CSEM, Switzerland \\
\tt\small{\{mshahbazi, entavelis, paudel, martin.danelljan, vangool\}@vision.ee.ethz.ch}, \and \tt\small{\{alessiot, edocollins\}@google.com}
}
\maketitle
\ificcvfinal\thispagestyle{empty}\fi

\begin{abstract}
Pose-conditioned convolutional generative models struggle with high-quality 3D-consistent image generation from single-view datasets, due to their lack of sufficient 3D priors. Recently, the integration of Neural Radiance Fields (NeRFs) and generative models, such as Generative Adversarial Networks (GANs), has transformed 3D-aware generation from single-view images. NeRF-GANs exploit the strong inductive bias of neural 3D representations and volumetric rendering at the cost of higher computational complexity. This study aims at revisiting pose-conditioned 2D GANs for efficient 3D-aware generation at inference time by distilling 3D knowledge from pretrained NeRF-GANs. We propose a simple and effective method, based on re-using the well-disentangled latent space of a pre-trained NeRF-GAN in a pose-conditioned convolutional network to directly generate 3D-consistent images corresponding to the underlying 3D representations. Experiments on several datasets demonstrate that the proposed method obtains results comparable with volumetric rendering in terms of quality and 3D consistency while benefiting from the computational advantage of convolutional networks. 
The code will be available at: \url{https://github.com/mshahbazi72/NeRF-GAN-Distillation}
\end{abstract}

\section{Introduction}
\label{sec:intro}

\begin{figure}[t]
\centering%
\includegraphics[width=0.94\linewidth]{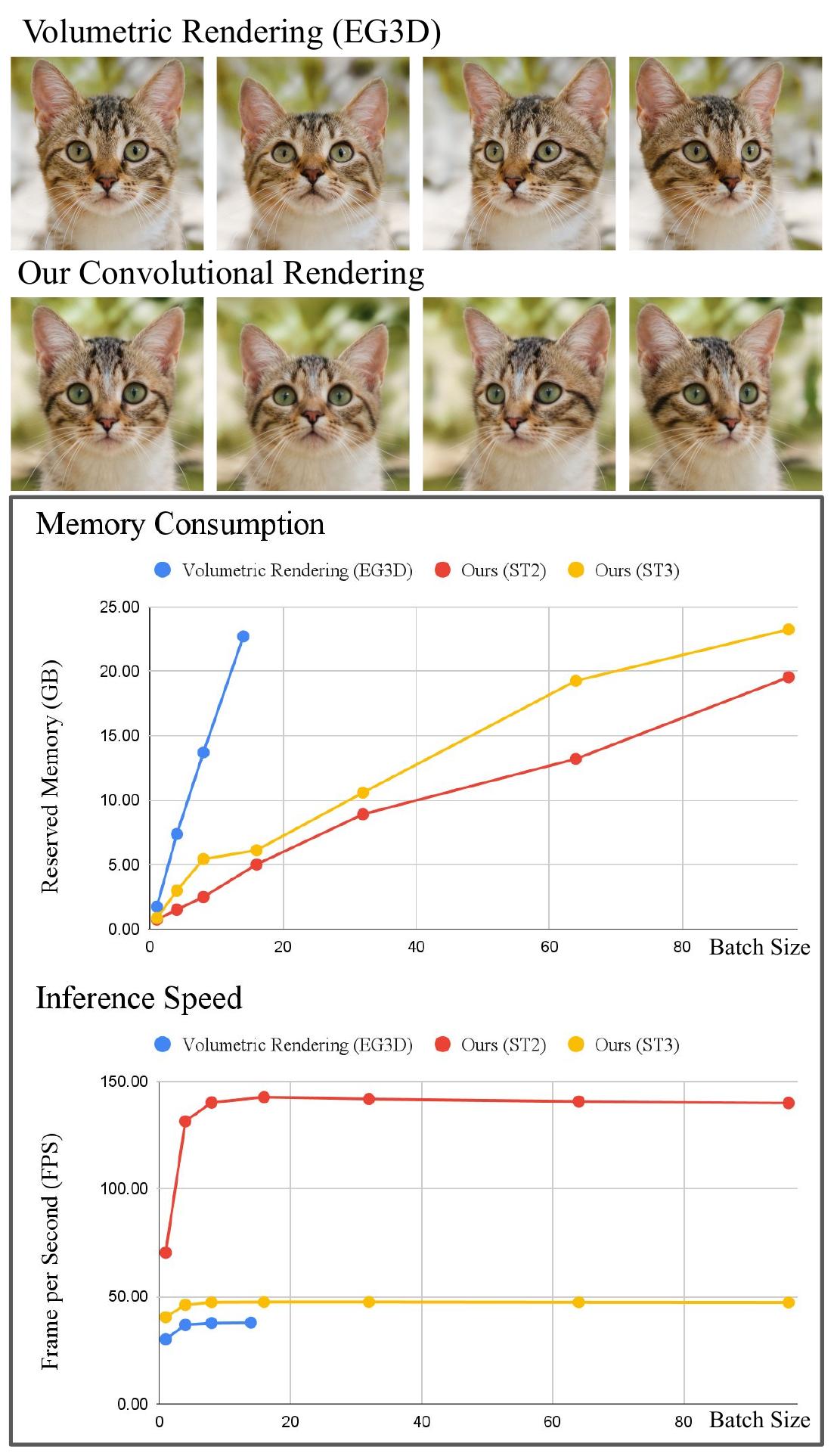}%
\vspace{-1.5mm}
\caption{Top: views of the same subject cat generated by a volumetric rendering generator (EG3D) and by our convolutional generator. Bottom: comparison of the inference memory consumption and speed (on a fixed GPU budget) for the two methods.}
\label{fig_main}%
\vspace{-4mm}
\end{figure}

Generative Adversarial Networks (GANs)~\cite{goodfellow2014gans} have undergone outstanding progress in photo-realistic image generation and manipulation in a variety of applications~\cite{brock2018large, Karras2019stylegan2, park2019SPADE, pix2pix2017, karras2018progressive, choi2020starganv2, Sauer2021ARXIV, Sauer2021ProjectedGC}.  Recently, there has been an increasing interest in extending GANs to the task of 3D-aware generation from single-view image datasets, with the goal of providing disentangled control over the content and the viewpoint of the generated images. 

Image GAN models have been historically based on convolutional architectures, enabling efficient training and generation for 2D tasks. However, pose-conditioned convolutional GANs (pcGANs) struggle with 3D-consistent image generation, due to their lack of sufficient 3D priors~\cite{HoloGAN2019}. Therefore, some studies have previously attempted to disentangle the pose from the content in pcGANs using explicit 3D supervision~\cite{rigstyle, KowalskiECCV2020Config, deng2020disco}, which, however, is not readily available for most datasets. As a result, later methods moved away from fully convolutional GANs by incorporating 3D inductive biases in the architecture and training pipeline, such as 3D neural representations and differentiable rendering methods~\cite{HoloGAN2019, BlockGAN2020, liftgan, pan2020gan2shape}.

The advent of Neural Radiance Fields (NeRFs)~\cite{mildenhall2020nerf} has recently transformed the neural 3D representation and the task of novel-view synthesis~\cite{nerftex, mipnerf, plenoxels, muller2022instant, nerfinthewild, nerfren, wu2021diver}. For this reason, NeRFs have been successfully integrated with GANs to achieve promising results in 3D-aware generation~\cite{Schwarz2020NEURIPS, chan2021pi, orel2022styleSDF, gu2021stylenerf, epigraf, niemeyer2021giraffe, schwarz2022voxgraf}. Nerf-GANs, in their general form, map a latent space to a 3D representation of objects and generate images from queried viewpoints using volumetric rendering. However, volumetric rendering is computationally demanding due to its ray-casting process, making high-resolution generation slow and memory-expensive. Recent works have proposed different approaches to improve the computational efficiency of NeRF-GANs using more efficient 3D representations~\cite{niemeyer2021giraffe, chan2021pi, schwarz2022voxgraf} and training protocols~\cite{chan2021pi, orel2022styleSDF}. Nevertheless, volumetric rendering remains an integral part of these models.

In recent NeRF-GANs~\cite{niemeyer2021giraffe, chan2021pi, schwarz2022voxgraf, orel2022styleSDF}, convolutional networks have been reintroduced in the generator architecture as super-resolution networks or as 3D-representation generators, in order to scale up NeRF-GANs for high-resolution generation. In this study, we take a different approach to integrating NeRF-GANs and convolutional GANs for 3D-aware generation from single-view images. In particular, we investigate the capacity of convolutional generators to achieve 3D-consistent rendering with explicit pose control when learning from a pretrained NeRF-GAN without any additional explicit 3D supervision. A convolutional generator that fairly preserves the 3D consistency, image quality, and the correspondence between the generated images and the underlying 3D representation can be used for efficient multi-view inference in setups where volumetric rendering is not affordable, such as in mobile applications. However, balancing and minimizing the trade-off between efficiency and 3D consistency is a highly challenging task, which we set out to explore in this work.

We propose a simple but effective method for distilling a pretrained NeRF-GAN into a pose-conditioned fully convolutional generator. The main component of our approach is based on exploiting the well-disentangled intermediate latent space of the NeRF-GAN in the convolutional generator. In particular, our convolutional generator learns to map each latent code from the 3D generator, along with the target viewpoint, to the corresponding obtained images by explicit volumetric rendering. By doing so, we aim to distill the NeRF-GAN's underlying 3D knowledge into the convolutional generator, as well as to establish a correspondence between the images of the generator and the 3D representation of the NeRF-GAN. As demonstrated in Fig.~\ref{fig_main}, our experiments on three different datasets indicate that the convolutional generator trained with our method is capable of achieving results  comparable to volumetric rendering in terms of image quality and 3D-consistency, while benefiting from the superior efficiency of convolutional networks. 
 
 Our contributions are summarized as follows:
\begin{itemize}[noitemsep]
  \item We propose a method to distill NeRF-GANs into convolutional generators for efficient 3D-aware inference.
  \item We provide a simple and effective method to condition the convolutional generator on the well-disentangled intermediate latent space of the NeRF-GAN.
  \item Through experiments on three different datasets, we show that the generator trained by our distillation method well preserves the 3D consistency, image quality, and semantics of the pretrained NeRF-GAN.
\end{itemize}

\section{Related Works}
\label{sec:related}

\noindent{\textbf{3D-aware Generation from Single-View Images.} Prior works have attempted to create 3D awareness in 2D GANs using explicit 3D supervision, such as 3D models~\cite{rigstyle, deng2020disco}, pose and landmark annotations~\cite{Shoshan_2021_gancontrol, hu2018rotation}, and synthetic data~\cite{KowalskiECCV2020Config}. In many applications, obtaining such 3D supervision is not practical. As a result, later works aimed at unsupervised methods by introducing 3D inductive biases in GANs, including 3D neural representations and differentiable rendering~\cite{HoloGAN2019, pan2020gan2shape, liftgan, BlockGAN2020} These methods, although promising, lag far behind 2D GANs in terms of image quality or struggle with high-resolution generation due to the additional computational complexity.

\noindent{\textbf{NeRF-GANs.}
NeRFS have shown outstanding potential in compactly representing 3D scenes for novel view synthesis. GRAF~\cite{Schwarz2020NEURIPS} and Pi-GAN~\cite{chan2021pi} are the first works to integrate NeRFs and GANs. While achieving highly consistent 3D-aware generation, the computational restrictions of NeRF framework make these methods impractical for high-resolution generation or environments with constrained resources. In order to extend NeRF-GANs to higher resolutions, convolutional super-resolution networks were used in later studies~\cite{niemeyer2021giraffe, gu2021stylenerf, orel2022styleSDF} at the expense of some multi-view inconsistencies. EpiGRAF~\cite{epigraf}, in contrast, adopts an efficient multi-scale patch training protocol, but still requires full high-resolution sampling rendering for inference, which makes it comparatively more computationally demanding than competitors.

Other studies aim at bringing the recent advances in the efficiency of NeRFS to NeRF-GANs. Although there exist numerous works on efficient 3D representations~\cite{mueller2022instant, plenoxels, sun2022direct, TensoRF, Wizadwongsa2021NeX, single_view_mpi} and volumetric sampling~\cite{yu2021plenoctrees, Garbin21arxiv_FastNeRF, hedman2021snerg, neff2021donerf, Hu_2022_CVPR} in NeRFs, only a subset of them~\cite{chan2022efficient, schwarz2022voxgraf, gmpi2022} have been successfully applied to NeRF-GANs. This is because they are mainly designed for the single-scene setup, making their adaptation to the generative setup not trivial. The use of sparse voxel grids in VoxGRAF~\cite{schwarz2022voxgraf} and multi-plane image representations in~\cite{gmpi2022} result in efficient and 3D-consistent generation while compromising the image quality and 3D geometry. EG3D~\cite{chan2022efficient} proposes using tri-planes to represent the geometry of the generated objects. Exploiting tri-planes, coupled with carefully designed techniques to enforce 3D consistency, allows EG3D to significantly improve both computational efficiency and image quality. Live 3D Portrait~\cite{trevithick2023} is a concurrent work based on EG3D that aims at real-time one-shot reconstruction of faces by estimating the canonical tri-planes of a pre-trained EG3D. However, Live 3D Portrait is computationally limited by the underlying volumetric rendering of EG3D.
Most similar to our study, SURF-GAN~\cite{kwak2022injecting} aims to discover directions for pose control in a pretraind 2D GAN by generating multi-view images using a pretrained NeRF-GAN. However, using NeRF-GANs only as multi-view supervision does not fully exploit their underlying 3D knowledge. Moreover, the 2D generator obtained using this method does not preserve any correspondence between the NeRF-GAN's 3D representations and the generated images. Different from SURF-GAN, we exploit the intermediate latent space of retrained NeRF-GANs to distill 3D knowledge into a 2D generator and establish correspondence between the convolutional generator and the NeRF-GAN's 3D representations.

\section{Method}
\label{sec:method}

\begin{figure*}[t]
    \centering%
    \includegraphics[width=0.95\linewidth]{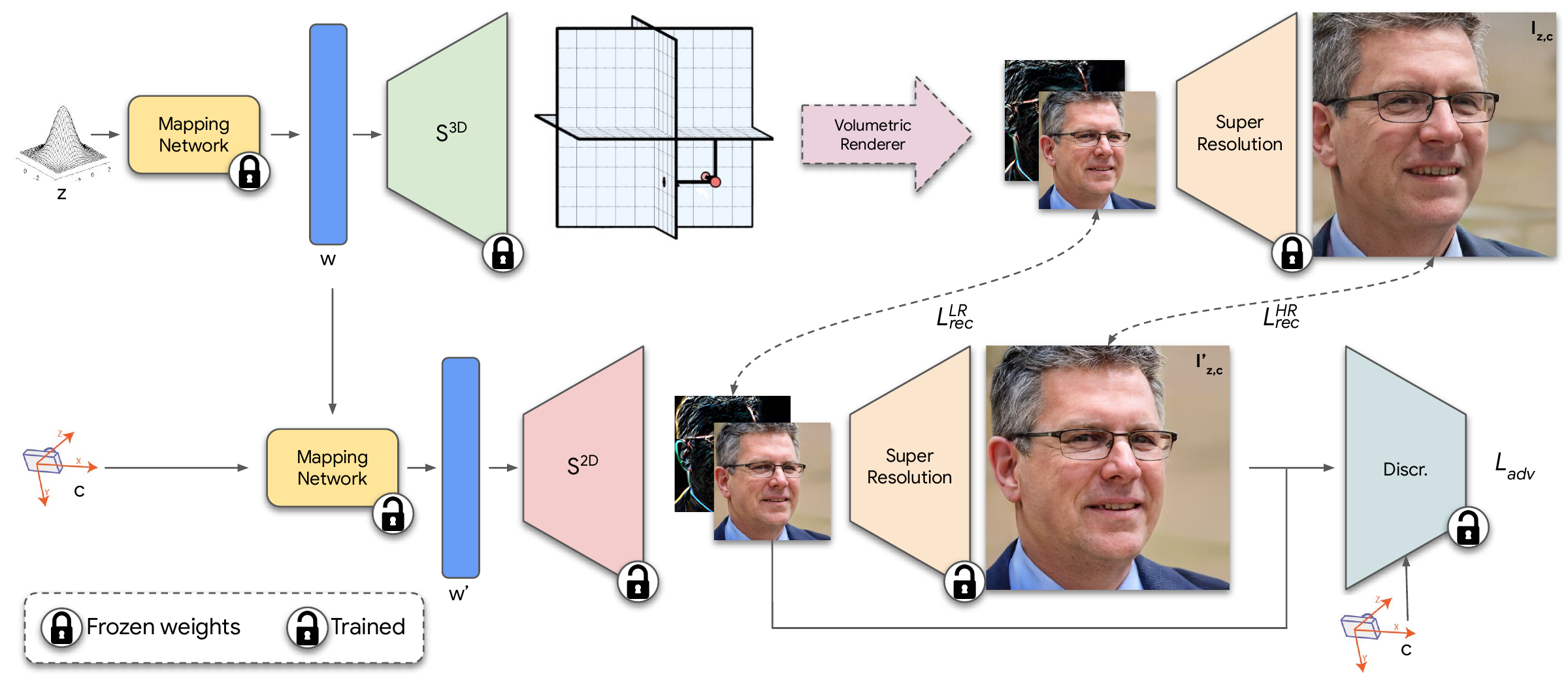}
    \vspace{-2mm}
    \caption{Using a student-teacher framework, we distill 3D consistency from a frozen volumetric rendering based Nerf-GAN (top) to a 2D convolutional renderer (bottom). A loss consisting of high- and low-resolution image reconstruction and an adversarial component allows us to retain good image quality and 3D consistency.}
    \label{fig_diagram}
\end{figure*}

In this section, we first provide a brief overview of the formulation of NeRF-GANs and then explain the proposed formulation in detail.

\subsection{Preliminaries}
\label{sec:preliminaries}
The general formulation of NeRF-GANs consists of a 3D-representation generator $G^{3D}(z)$, which maps a latent variable $z$ (usually drawn from a normal distribution) to a 3D representation of an object. Then, in order to render an image $I_{z,c}$ from the target viewpoint (camera parameters) $c \in R^{25}$, volumetric rendering is applied to the generated 3D representation. We base our method on EG3D~\cite{chan2022efficient}, as it provides a strong trade-off in image quality, 3D consistency, and efficiency, among recent NeRF-GANs.

EG3D represents 3D scenes using tri-planes, which are three axis-aligned orthogonal feature planes, each with a size of $N \times N \times C$, where $N$ is spatial resolution and $C$ is the number of channels. To represent a 3D position $x \in R^3$, $x$ is projected onto each of the three feature planes, retrieving the corresponding feature vector $(F_{xy}, F_{xz}, F_{yz})$ via bilinear interpolation, and aggregating the three feature vectors via summation. To obtain the color and density at position $x$, a lightweight MLP decodes the feature vector obtained for the queried position to a density and color value.

The tri-plane generator $G^{3D}(z, c)$ in EG3D consists of a mapping network $M^{3D}(z, c)$, which maps the input latent code and the target viewpoint to an intermediate latent variable $w$, namely the style code. The style code then is used to modulate a convolutional synthesis network $S^{3D}(w)$ to generate the tri-planes $I^{3p}_{z,c}$. In order to render an image $I_{z,c}$ from the target viewpoint $c$, hierarchical volumetric rendering is applied to the tri-planes. Since volumetric rendering at high resolutions is computationally too expensive, EG3D does so at a lower resolution and uses a convolutional super-resolution network to obtain a final image. More specifically, the low-resolution output of volumetric rendering in EG3D consists of a 32-channel feature map $I^f_{z,c}$, the first three of which represent the low-resolution RGB image $I^{LR}_{z,c}$), which is given as input to the super-resolution network. EG3D is trained in an adversarial fashion with a viewpoint-conditioned dual discriminator $D$ that ensures the photorealism of the generated images from the target viewpoints, as well as the consistency between high-resolution and low-resolution images.

\subsection{Convolutional Rendering of Pretrained NeRF-GANs} 

The aim of the proposed method is to distill a pre-trained NeRF-GAN $G^{3D}_{z,c}$ into a 2D image generator $G^{2D}_{z, c}$, such that $G^{2D}_{z, c}$ directly predicts 3D-consistent multi-view images $I'_{z, c}$, corresponding to the volumetric renderings obtained by the underlying 3D representation of $G^{3D}_{z,c}$. To this end, we propose to exploit the well-disentangled style space of $G^{3D}$ to distill the underlying 3D representation into $G^{2D}$. Sharing the style space $w$ of the pre-trained 3D generator with the convolutional renderer is the first step towards establishing a correspondence between the 3D representations and the generated images. Secondly, it allows training the convolutional generator for 3D-consistent generation without the need for generating multiple views of the same objects and enforcing multi-view consistency. 

The overall architecture of EG3D and our convolutional generator is visualized in \ref{fig_diagram}. The convolutional generator is based on StyleGAN architecture~\cite{Karras2020ada, karras2021alias}, consisting of a mapping network, a low-resolution convolutional feature prediction, and a convolutional super-resolution network. The mapping network transforms the style code $w$ of $G^{3D}$, and the target viewpoint $c$ to the style code $w'$ of $G^{2D}$. Then, the low-resolution feature predictor $S^{2D}$ estimates the EG3D features and low-resolution image obtained by the volumetric rendering. Estimated features and images are then mapped to the high-resolution outputs using the super-resolution network. In our setup, the super-resolution network is initialized with EG3D's super-resolution network and is jointly optimized with the feature predictor network.

\subsection{Training}
\label{sec:training}

In order to train the proposed convolutional renderer, we use a teacher-student framework, where the volumetric rendering is used to supervise $G^{2D}$ on the viewpoint-conditioned mapping of $(z, c)$ to $I'_{z,c}$.
A schematic representation of our training regime is reported in Fig.~\ref{fig_diagram}.
Specifically, for each training sample, we randomly sample $z_i$ and $c_i$ and use the pretrained NeRF-GAN to obtain: the corresponding style code $w_i$, the low-resolution image $I^{LR}_{z_i,c_i}$, feature maps $I^{f}_{z_i,c_i}$ rendered by volumetric rendering, as well as, the high-resolution image $I^{HR}_{z_i,c_i}$ generated by the super-resolution network. These together form a training sample $i$ for the proposed convolutional renderer.

We provide $z$ and $c$ to $G^{2D}$ and compute a loss function composed of three parts. We first add a reconstruction term $L^{LR}_{rec}$ between the low-resolution outputs of the volumetric and convolutional renderers. A second reconstruction loss $L^{HR}_{rec}$ is applied between the super-resolved outputs of the two renderers. Lastly, we apply an adversarial term $L_{adv}$.

In the following, we drop the subscripts $z_i,c_i$ to reduce the clutter in notation. The low-resolution reconstruction loss $L^{LR}_{rec}$ consists of a pixel-wise smooth L1 loss between the two feature maps, as well as a perceptual loss between the generated and target low-resolution images,
\begin{equation}
\label{eq:loss_lr}
    \begin{split}
        L^{LR}_{rec} = &\lambda^{LR}_{\text{Smooth}L1} * \text{Smooth}L1(I'^{f}, I^{f}) + \\
        &\lambda^{LR}_{\text{perc}} * \text{PerceptualLoss}(I'^{LR}, I^{LR}).
    \end{split}
\end{equation}
Here, $\lambda^{LR}_{\text{Smooth}L1}$ and $\lambda^{LR}_{perc}$ are the weights for the low-resolution smooth L1 and perceptual loss, respectively.

Similarly, the high-resolution reconstruction loss $L^{HR}_{rec}$is defined as:
\begin{equation}
\label{eq:loss_hr}
    \begin{split}
         L^{HR}_{rec} = &\lambda^{HR}_{\text{Smooth}L1} * \text{Smooth}L1(I'^{f}, I^{f}) + \\
        &\lambda^{HR}_{\text{perc}} * \text{PerceptualLoss}(I'^{HR}, I^{HR}).
    \end{split}
\end{equation}
where $\lambda^{HR}_{rec}$ and $\lambda^{HR}_{perc}$ are the weights for the high-resolution smooth L1 and perceptual loss, respectively.

The adversarial term $L_{adv}$ is similar to the one used in EG3D. We use the same dual-discriminator architecture $D$ as in EG3D to ensure the realism of the high-resolution images, their consistency with the low-resolution version, and the compliance of the generated image with the queried viewpoints. The total loss $L_{total}$ for training $G^{2D}$ is,
\begin{equation}
    \label{eq:loss}
    L_{total} = L^{LR}_{rec} + L^{HR}_{rec} + \lambda_{adv} *  L_{adv},
\end{equation}
where $\lambda_{adv}$ is the weight for the adversarial loss.

\textbf{Two-stage training:} in practice, empirical experiments show that training the convolutional renderer using the full objective from the beginning will lead to high-quality but 3D-inconsistent images. Therefore, we instead propose a 2-stage training curriculum. In the first stage, $G^{2D}$ is only optimized by pure distillation of the volumetric rendering $G^{3D}$ using $L^{LR}_{rec}$ and $L^{HR}_{rec}$ until the renderer achieves reasonable generation quality. Then, the adversarial loss $L_{adv}$ is added to the training to further improve the performance. By applying this 2-stage curriculum, we are able to  counter the 3D inconsistency induced by the adversarial training.

\textbf{Pose-correlated dataset bias:} As shown in EG3D~\cite{chan2022efficient}, adversarial training of the convolutional network is prone to learning pose-correlated dataset biases, such as more smiling in non-frontal viewpoints in FFHQ dataset~\cite{Karrasstyle}, which in turn results in 3D attribute inconsistencies. To mitigate such biases in FFHQ dataset, we use both real images and the images rendered from EG3D as the real examples shown to the discriminator. The proportion of the EG3D-rendered images shown to $D$ is controlled by the hyper-parameter $\alpha~(0\le\alpha\le1)$, which is set to $0.5$ in our experiments. As we will discuss in section~\ref{sec_ablation}, $\alpha$ can be used to control the trade-off between image quality and 3D consistency.

\section{Experiments}

In this section, we first describe our experimental setup for the evaluation of our method. Then, we compare the proposed method with baselines in terms of visual quality, 3D consistency, and computational efficiency. Moreover, we provide an ablation study and a discussion on the benefits and trade-offs of the proposed method.

\subsection{Datasets}\label{sec:dataset}
Following EG3D~\cite{chan2021pi}, we evaluate our method on three different datasets:

\noindent\textbf{Flickr-Faces-HQ (FFHQ)~\cite{Karrasstyle}}: a collection of 70k high-quality images of real-world human faces, as well as corresponding approximate camera extrinsics estimated using an of-the-shell pose estimator.

\noindent\textbf{AFHQ Cats}: a sub-category of the Animal-Face-HQ (AFHQ)~\cite{choi2020stargan}, consisting of around 5k high-quality images of cat faces, as well as corresponding camera extrinsics estimated using an of-the-shell pose estimator.

\noindent\textbf{ShapeNet Cars:} a category of ShapeNet~\cite{chang2015shapenet} consisting of synthetic images of cars rendered from different viewpoints, as well as the corresponding camera extrinsics annotations.

\subsection{Baselines}\label{sec:baseline}
We consider EG3D~\cite{chan2022efficient} and The method proposed in SURF-GAN~\cite{kwak2022injecting} as our main baselines for this study. For a more complete evaluation, we also compare our method to additional relevant baselines:

\noindent\textbf{EG3D~\cite{chan2022efficient}}: the NeRF-GAN used for distilling 3D knowledge in the convolutional generator. EG3D serves as the upper bound for the 3D consistency of the proposed method.

\noindent\textbf{Pose-Conditioned StyleGAN (PC-GAN)}: a standard conditional 2D GAN, conditioned on the pose annotations without any knowledge distillation.

\noindent\textbf{SURF}: Inspired by the proposed method in SURF-GAN~\cite{kwak2022injecting}, we create a baseline called SURF, where multi-view images of EG3D are used to discover pose-control in a 2D StyleGAN (pretrained on FFHQ) based on the proposed method in~\cite{kwak2022injecting}. The details of the implementation can be found in the supplementary material\footnote{As the official implementation of SURF-GAN does not include their method for pose control in 2D GANs, we provide our best attempt in implementing their method.}.

\noindent\textbf{LiftGAN~\cite{liftgan}}: a method predating EG3D and SURF baselines, based on differentiable rendering for distilling 2D GANs to train a 3D generator.

\subsection{Implementation and Evaluation Details}\label{sec:detail}
We implement and evaluate the proposed generator using both StyleGAN2 (ST2)~\cite{Karras2020ada} and StyleGAN3 (ST3)~\cite{karras2021alias} architectures. For the pretrained NeRF-GAN, we use the official models from EG3D~\cite{chan2022efficient} (for Shapenet Cars, we re-train the model as the official model does not match the results reported by EG3D). We train our experiments using a batch size of 16. The rendering resolution and the final resolution are (128, 512) for FFHQ and AFHQ and (64, 128) for Shapenet Cars. Both training and inference experiments were conducted using NVidia RTX 3090 GPUs. In all experiments, we set all of the weights of reconstruction loss terms ($\lambda^{LR}_{smooth_l1}$, $\lambda^{LR}_{perc}$, $\lambda^{HR}_{perc}$) to the value 1 and the weight of the adversarial loss ($\lambda_{adv}$) to the value 0.1.

\subsection{Metrics}\label{sec:metric}
We evaluate our method quantitatively in terms of visual quality, and 3D consistency.

\noindent\textbf{Fréchet Inception Distance (FID)~\cite{heusel2017fid}}: The most common metric to assess the quality and diversity of generation.

\noindent\textbf{Kernel Inception Distance (KID)~\cite{heusel2017fid}}: An unbiased alternative to FID for smaller datasets.

\noindent\textbf{Pose Accuracy (PA)}: following previous works~\cite{chan2022efficient}, we measure the ability of the model in generating images of the query poses by calculating the mean squared error (MSE) between the query poses and the pose of the generated images, estimated using an of-the-shelf pose estimator~\cite{deng2019accurate}.

\noindent\textbf{Identity Preservation (ID)}: As a metric for 3D consistency, we measure the degree of face identity preservation between different viewpoints with respect to the canonical pose using ArcFace~\cite{deng2018arcface} cosine similarity for the FFHQ setup.

\noindent\textbf{3D Landmark Consistency}: As another 3D consistency metric, we measure the change in facial landmarks between different viewpoints in FFHQ using MSE. The 3D landmarks are estimated using an off-the-shelf estimator~\cite{deng2019accurate}.

\subsection{Quantitative Comparison}
In the following, we quantitatively compare the proposed method with the baselines described in Sec.~\ref{sec:baseline} in terms of inference efficiency, visual quality, and 3D consistency.

\subsubsection{Efficiency}
The efficiency of fully-convolutional networks compared to the rendering-based method is well-known. To better assess the practical benefit of the proposed method, we provide a comparison of inference efficiency between EG3D. Fig~\ref{fig_main} visualizes an example of the inference memory consumption and speed of the two methods using different batch sizes on a fixed GPU budget (in this case, on RTX 3090 GPU with 24G of memory). 
As shown, EG3D is restricted to small batch sizes (a maximum of 14) due to its costly memory consumption, where our method can scale up to a maximum of 96 samples per batch.
As for speed, our convolutional generator achieves better frame-per-second, especially when using StyleGAN2 as its backbone.

\subsubsection{Image Quality}
To assess the trade-off brought about by our convolutional generator, we evaluate the quality of the generated images. Table~\ref{tab_fid} shows the FID and KID scores for our method and the baseliness on different datasets. Compared to the PC-GAN and SURF baselines, our method constantly achieves higher quality. This confirms that exploiting the style space of the pretrained NeRF-GAN contributes to the ability of the convolutional renderer in pose-conditioned generation. Although our method does not fully match the visual quality of EG3D, it is still able to fairly maintain high image quality and significantly reduce the compromise in the quality compared to the other convolutional counterparts.
\begin{table}[t]
\caption{Comparison of image quality on three datasets in terms of FID and KID metrics. *The value is borrowed from~\cite{liftgan}.}%
\label{tab_fid}%
\centering%
\resizebox{0.98\linewidth}{!}{%
\begin{tabular}{lllllll}
\toprule
\multicolumn{1}{l}{\bf Method} &\multicolumn{2}{c}{\bf FFHQ} &\multicolumn{2}{c}{\bf AFHQ} &\multicolumn{2}{c}{\bf ShapeNET Cars} \\ 
\multicolumn{1}{l}{} &\multicolumn{1}{c}{FID $\downarrow$} &\multicolumn{1}{c}{KID $\downarrow$} &\multicolumn{1}{c}{FID $\downarrow$} &\multicolumn{1}{c}{KID $\downarrow$} &\multicolumn{1}{c}{FID $\downarrow$} &\multicolumn{1}{c}{KID $\downarrow$} \\
\midrule
\multicolumn{1}{l}{EG3D~\cite{chan2022efficient}} &\multicolumn{1}{c}{5.0} &\multicolumn{1}{c}{0.0018} &\multicolumn{1}{c}{2.9} &\multicolumn{1}{c}{0.0003} &\multicolumn{1}{c}{3.5} &\multicolumn{1}{c}{0.0017} \\
\midrule
\multicolumn{1}{l}{PC-GAN}  &\multicolumn{1}{c}{19.3} &\multicolumn{1}{c}{0.0085} &\multicolumn{1}{c}{4.5} &\multicolumn{1}{c}{0.0009} &\multicolumn{1}{c}{6.1} &\multicolumn{1}{c}{0.0018} \\
\multicolumn{1}{l}{LiftGAN~\cite{liftgan}}  &\multicolumn{1}{c}{29.8*} &\multicolumn{1}{c}{-}  &\multicolumn{1}{c}{-} &\multicolumn{1}{c}{-} &\multicolumn{1}{c}{-} &\multicolumn{1}{c}{-} \\
\multicolumn{1}{l}{SURF}  &\multicolumn{1}{c}{31.1} &\multicolumn{1}{c}{0.0153} &\multicolumn{1}{c}{-} &\multicolumn{1}{c}{-} &\multicolumn{1}{c}{-} &\multicolumn{1}{c}{-} \\
\multicolumn{1}{l}{Ours (ST2)}  &\multicolumn{1}{c}{6.6} &\multicolumn{1}{c}{0.0019} &\multicolumn{1}{c}{3.8} &\multicolumn{1}{c}{0.0011} &\multicolumn{1}{c}{3.1} &\multicolumn{1}{c}{0.0013} \\
\multicolumn{1}{l}{Ours (ST3)}  &\multicolumn{1}{c}{6.8} &\multicolumn{1}{c}{0.0023} &\multicolumn{1}{c}{3.2} &\multicolumn{1}{c}{0.0007} &\multicolumn{1}{c}{3.1} &\multicolumn{1}{c}{0.0012} \\
\bottomrule
\end{tabular}}
\vspace{-3mm}
\end{table}

\subsubsection{3D Consistency}
While, unlike volumetric rendering, 2D convolutions do not guarantee 3D consistency, we show that our approach achieves a good performance in this regard. We assess the 3D consistency of generated images on FFHQ by measuring the pose accuracy, 3D landmark consistency, and face identity preservation, as discussed in Sec. \ref{sec:metric}, which are provided in Table \ref{tab_3d}. Based on the results, Our method achieves comparable 3D consistency with EG3D, while PC-GAN and SURF struggle. Note that the high values for identity preservation and 3D landmark consistency in SURF are due to the limited pose variations, and hence generating similar images regardless of the input pose, as reflected by the pose accuracy (and the visual examples in Fig.~\ref{fig_vis_ffhq}). 
\begin{table}[t]
    \caption{Comparison of 3D consistency metrics on FFHQ.}
    \label{tab_3d}%
    \centering
        \resizebox{0.75\linewidth}{!}{%
        \begin{tabular}{lllll}
        \toprule
        \multicolumn{1}{l}{\bf Method} &\multicolumn{1}{c}{\bf Pose Acc. $\downarrow$} &\multicolumn{1}{c}{\bf 3D Landmark $\downarrow$} &\multicolumn{1}{c}{\bf ID $\uparrow$} \\ 
        \midrule
        \multicolumn{1}{l}{EG3D~\cite{chan2022efficient}} &\multicolumn{1}{c}{0.002} &\multicolumn{1}{c}{0.018} &\multicolumn{1}{c}{0.75}\\
        \midrule
        \multicolumn{1}{l}{PC-GAN} &\multicolumn{1}{c}{0.009} &\multicolumn{1}{c}{0.062} &\multicolumn{1}{c}{0.56}\\
        \multicolumn{1}{l}{SURF} &\multicolumn{1}{c}{0.044} &\multicolumn{1}{c}{0.014} &\multicolumn{1}{c}{0.86}\\
        \multicolumn{1}{l}{Ours (ST2)} &\multicolumn{1}{c}{0.002} &\multicolumn{1}{c}{0.023}  &\multicolumn{1}{c}{0.75}\\
        \multicolumn{1}{l}{Ours (ST3)} &\multicolumn{1}{c}{0.002} &\multicolumn{1}{c}{0.022}  &\multicolumn{1}{c}{0.75}\\
        \midrule
        \end{tabular}} 
        \vspace{-5mm}

\end{table}

\begin{figure*}
\begin{center}
\includegraphics[width=0.92\linewidth]{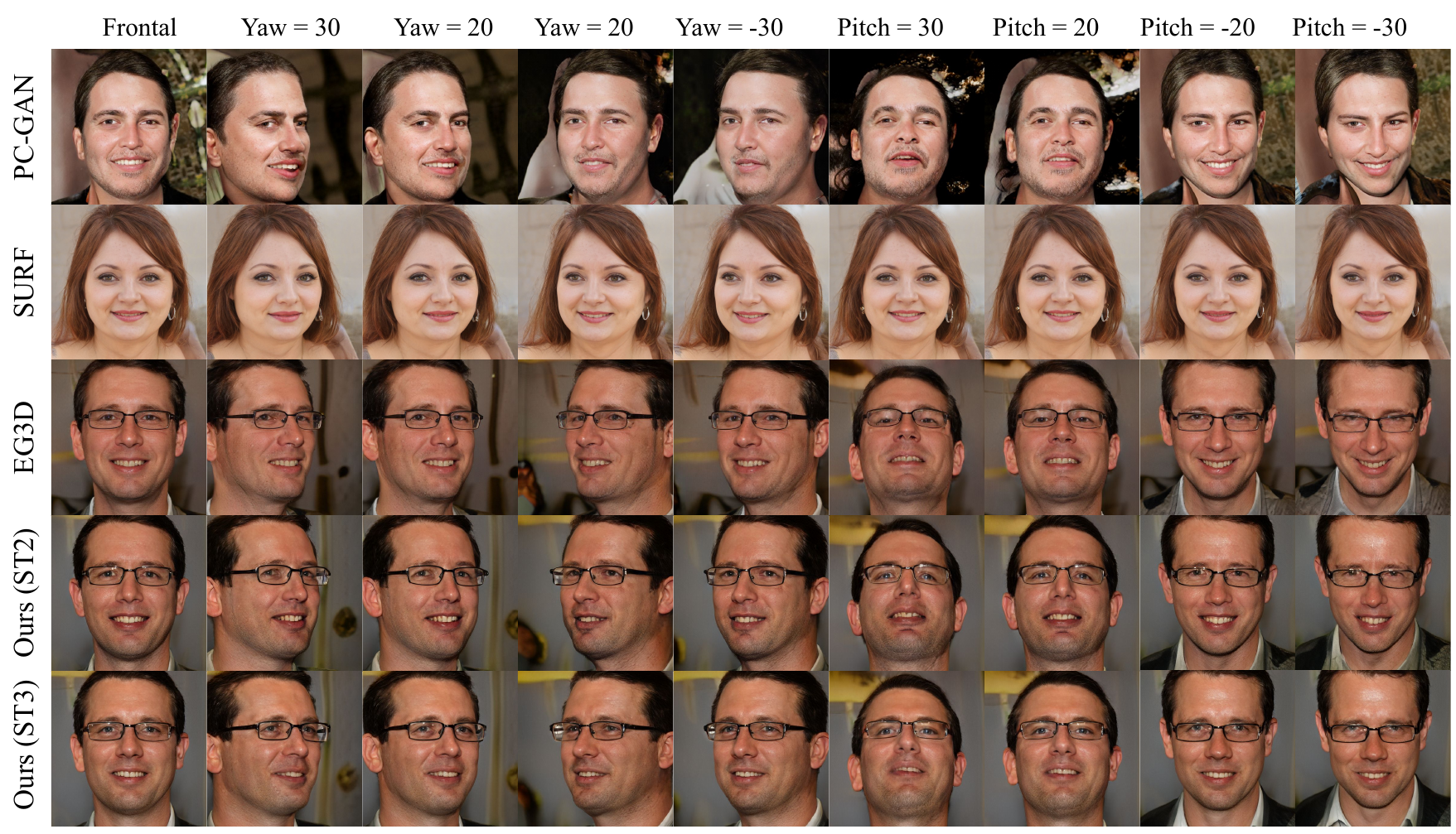}
\end{center}
\vspace{-4mm}
\caption{Qualitative examples of variations in yaw and pitch for FFHQ. Compared to the pose-conditioned GAN and SURF baseline, our proposed method nearly matches the 3D consistency and image quality of volumetric rendering (EG3D).}
\label{fig_vis_ffhq}
\vspace{-3mm}
\end{figure*}

\begin{figure*}
\begin{center}
\includegraphics[, width=0.92\linewidth]{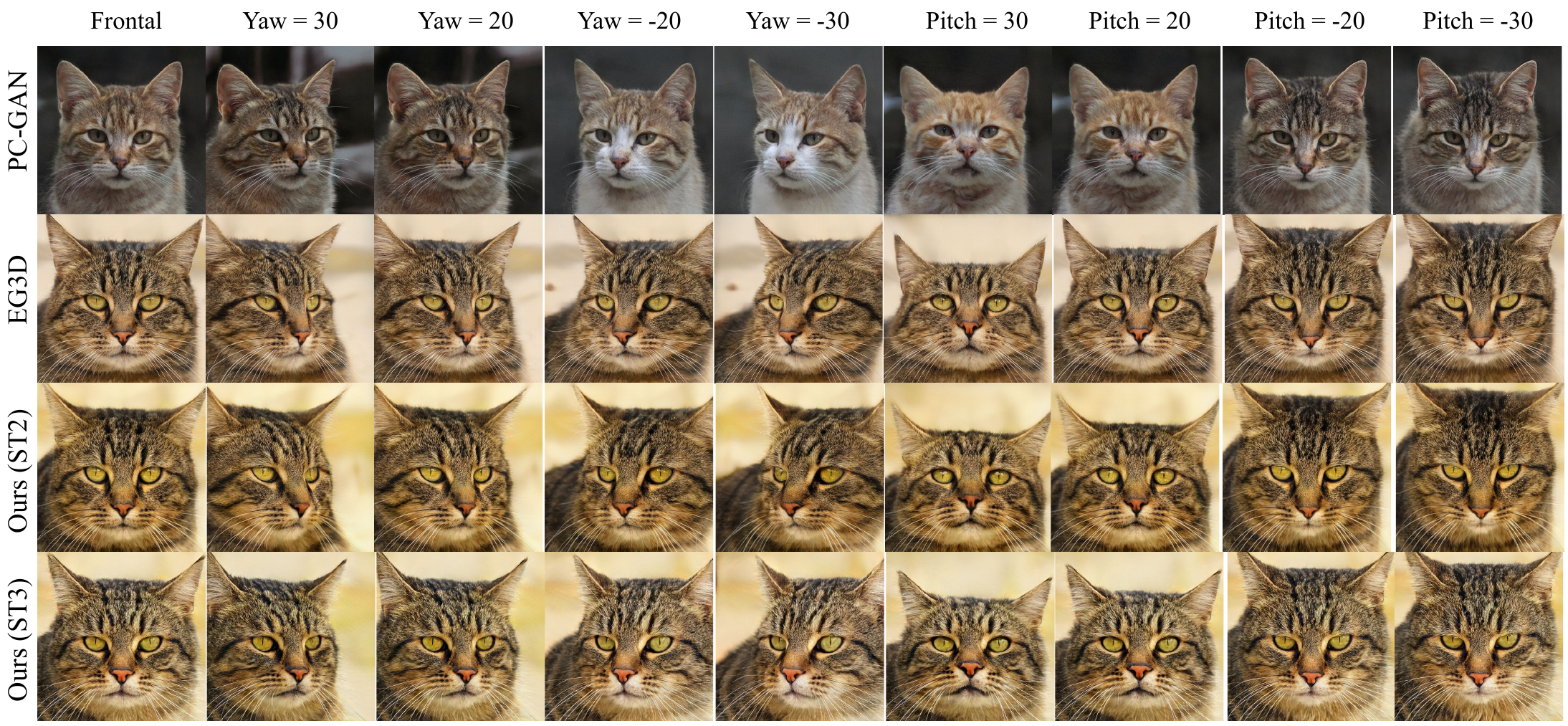}
\end{center}
\vspace{-4mm}
\caption{Qualitative examples of variations in yaw and pitch for AFHQ cats. In line with our quantitative experiments, the pose-conditioned convolutional baseline (PC-GAN) fails to preserve the identity of the subject under different poses. In contrast, our method exhibits similar preservation of identity to the volume rendering approach (EG3D), despite the difference in computational resources and time.}
\label{fig_vis_cats}
\vspace{-4mm}
\end{figure*}

\begin{figure*}
\begin{center}
\includegraphics[width=\linewidth]{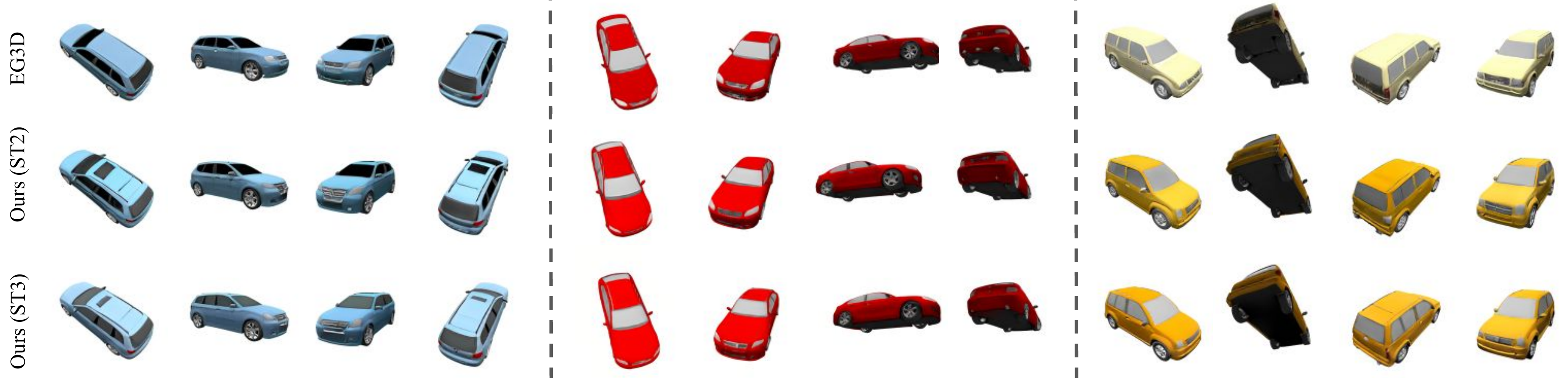}
\end{center}
\vspace{-4mm}
\caption{Qualitative examples of different camera poses in Shapenet Cars for three different car models.}
\label{fig_vis_cars}
\vspace{-4mm}
\end{figure*}

\subsection{Ablation}\label{sec_ablation}
\noindent\textbf{Ablation on loss functions:} the proposed training objective in section~\ref{sec:training} consists of different loss terms to ensure both image quality and consistency with the output of volumetric rendering. In this section, we ablate the importance of these components, by adding them one by one to form the final objective of Eq.~\ref{eq:loss}. Table~\ref{tab_ablation} shows the FID scores for the following experiments on the loss terms on AFHQ dataset:
\begin{itemize}[noitemsep]
  \item \textbf{LR}: Only the low-resolution reconstruction loss. the super-resolution network is frozen.
  \item \textbf{HR}: Only the high-resolution reconstruction loss.
  \item \textbf{LR + HR} Full reconstruction loss on low-resolution and high-resolution.
  \item \textbf{HR + Adv}: reconstruction and adversarial losses on the high-resolution images.
  \item \textbf{Full (LR + HR + ADV)} Full training objective, including the reconstruction and adversarial terms.
\end{itemize}
As shown by the ablation study, the combination of the proposed loss terms leads to the best FID scores.

\begin{table}[b]
\vspace{-3mm}
\caption{Ablation on different loss functions for training the convolutional renderer on AFHQ Cats dataset.}%
\label{tab_ablation}%
\centering%
\resizebox{0.6\linewidth}{!}{%
\begin{tabular}{ll}
\toprule
\multicolumn{1}{l}{\bf Method} &\multicolumn{1}{c}{\bf FID $\downarrow$} \\ 
\midrule
\multicolumn{1}{l}{LR} &\multicolumn{1}{c}{30.55}   \\
\multicolumn{1}{l}{HR} &\multicolumn{1}{c}{10.39}  \\
\multicolumn{1}{l}{LR + HR} &\multicolumn{1}{c}{9.1}   \\
\multicolumn{1}{l}{HR + ADV} &\multicolumn{1}{c}{6.58}   \\
\multicolumn{1}{l}{Full (LR + HR + ADV)} &\multicolumn{1}{c}{3.2}   \\
\bottomrule
\end{tabular}}
\vspace{-2mm}
\end{table}

\begin{table}[b]
\caption{The effect of mixing real images and EG3D-rendered images as real examples for adversarial training, controlled by the parameter $\alpha$, on FFHQ dataset.}%
\label{tab_alpha}%
    \centering
    \resizebox{\linewidth}{!}{%
    \begin{tabular}{lllll}
    \toprule
    \multicolumn{1}{l}{\bf Method} &\multicolumn{2}{c}{\bf Ours (ST2)} &\multicolumn{2}{c}{\bf Ours (ST3)}\\ 
    \multicolumn{1}{l}{} &\multicolumn{1}{c}{FID $\downarrow$} &\multicolumn{1}{c}{3D Landmark $\downarrow$} &\multicolumn{1}{c}{FID $\downarrow$} &\multicolumn{1}{c}{3D Landmark $\downarrow$} \\
    \midrule
    \multicolumn{1}{l}{$\alpha=0$}  &\multicolumn{1}{c}{5.5} &\multicolumn{1}{c}{0.027} &\multicolumn{1}{c}{5.7} &\multicolumn{1}{c}{0.027} \\
    \multicolumn{1}{l}{$\alpha=0.5$}  &\multicolumn{1}{c}{6.8} &\multicolumn{1}{c}{0.022} &\multicolumn{1}{c}{6.6} &\multicolumn{1}{c}{0.023}\\
    \bottomrule
    \end{tabular}}
\end{table}

\noindent\textbf{Single-stage v.s. two-stage training:} As mentioned in \ref{sec:training}, we find out that single-stage training by jointly optimizing for both reconstruction and adversarial losses results in subtle inconsistencies such as color shifts and geometry warps, which can be mitigated using the proposed 2-stage training in section~\ref{sec:method}. As the observed inconsistencies are difficult to capture using our quantitative 3D consistency metrics, we provide a visual comparison between the examples of single-stage and two-stage training on AFHQ in Fig. \ref{fig_stage}. To better visualize the inconsistencies, we provide a video visualization and comparison in the supplementary material.

\noindent\textbf{Mitigating Pose-Attribute Correlation:} In Table~\ref{tab_alpha}, we provide an ablation on the parameter $\alpha$ introduced in section~\ref{sec:training} for FFHQ dataset. As shown, including EG3D-generated images ($\alpha=0.5$) improves the 3D consistency at the cost of lower generation quality.

\begin{figure}
\begin{center}
\includegraphics[width=\linewidth]{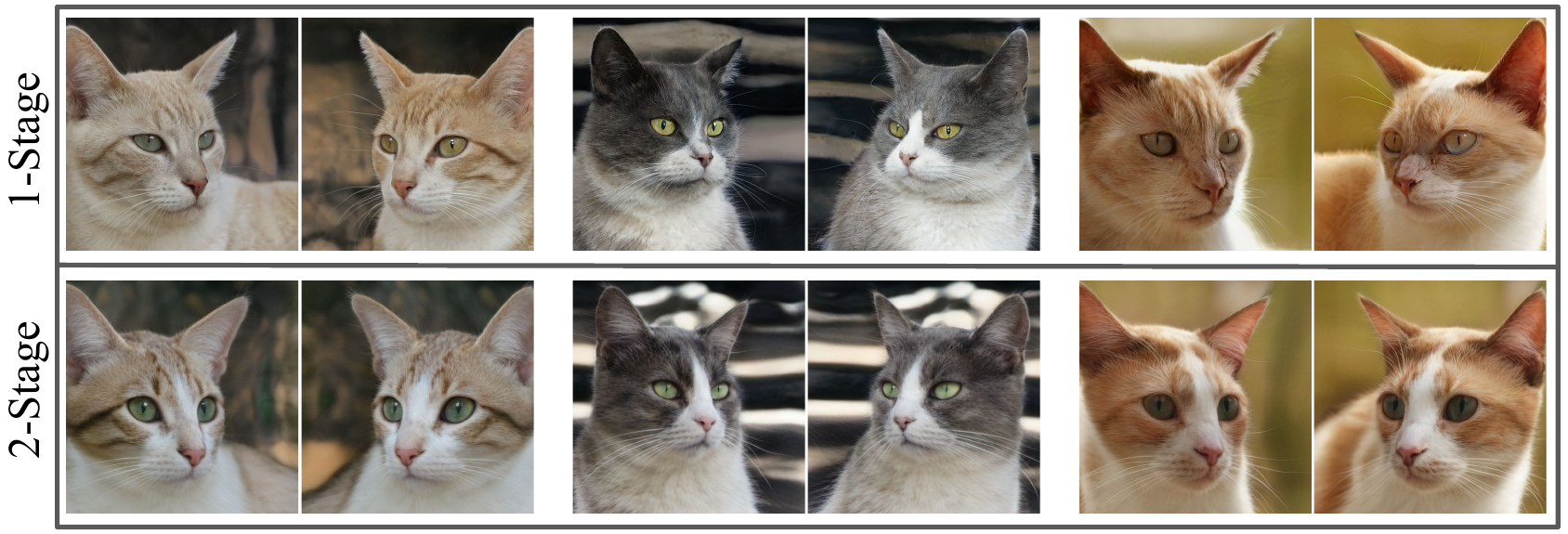}
\end{center}
\vspace{-3mm}
\caption{one-stage training causes subtle color and geometry inconsistencies (first row). Such inconsistencies can be resolved using our proposed 2-stage training (second row).}
\vspace{-3mm}
\label{fig_stage}
\vspace{-1mm}
\end{figure}

\subsection{Qualitative Comparison}

In this section, We provide a visual comparison of our method with the baselines. In Fig.~\ref{fig_vis_ffhq} and ~\ref{fig_vis_cats}, we provide visual examples of variations in yaw and pitch for FFHQ and AFHQ Cats. Compared to a PC-GAN and SURF, our proposed method closely matches the 3D consistency and maintains the image quality of volumetric rendering. Fig.~\ref{fig_vis_cars} additionally provides examples of Shapenet Cars generated using our method and their corresponding images from EG3D. Similarly, our method exhibits preservation of 3D consistency and image quality, despite the difference in required computational resources. We provide more visual results in the supplementary material.

\subsection{Inversion, Interpolation, and Style Mixing}

As the proposed generator follows a StyleGAN architecture, it can easily benefit from most of the editing techniques common in the GANs' literature. Fig.~\ref{fig_edit} shows examples of inversion using Pivotal Tuning Inversion (PTI)~\cite{roich2022pivotal}, latent space interpolation, and style mixing. 

\begin{figure}
\begin{center}
\includegraphics[width=\linewidth]{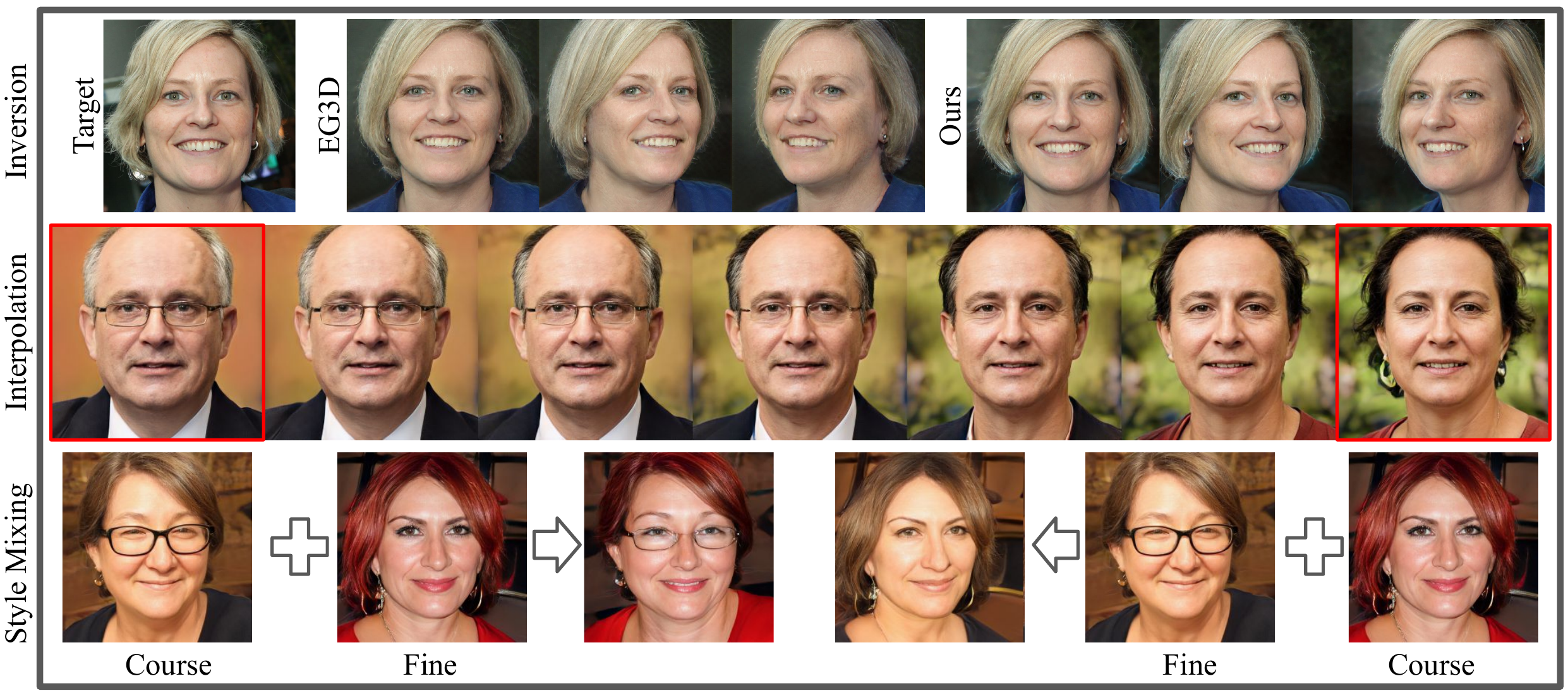}
\end{center}
\vspace{-4mm}
\caption{First row: Inversion using PTI ~\cite{roich2022pivotal}) for EG3D and our method; second row: interpolation in the latent space of our method; third row: style mixing in the latent space of our method.}
\label{fig_edit}
\vspace{-4mm}
\end{figure}

\subsection{Discussion: StyleGAN2 V.S. StyleGAN3}\label{arch}
StyleGAN2 is more computationally efficient than StyleGAN3. Based on the provided quantitative evaluations, our method reaches comparable image quality and 3D consistency with both architectures. However, StyleGAN2 is known to suffer from more texture stitching and artifacts~\cite{karras2021alias}, which we also observe in the generated images (visualizations are provided in the supplementary material).

\subsection{Correspondence between Convolutional and Volumetric Renderering}
As mentioned before, exploiting the style space of the pretrained NeRF-GAN also provides an opportunity for establishing a direct correspondence between the 3D representation of the 3D generator and the generated images using the convolutional generator. A close comparison of images generated using the convolutional and volumetric rendering in Figures~\ref{fig_vis_ffhq}, ~\ref{fig_vis_cats}, and, ~\ref{fig_vis_cars} indicates that the convolutional render is able to infer and match many attributes of the underlying 3D representation from the shared latent space and generate images similar in content to those of volumetric rendering. However, there still remains a gap in the full correspondence of the two rendering methods, as semantic and identity changes are visible between the corresponding images generated by the two methods. Investigating more explicit approaches for enforcing correspondence could be an interesting direction for improving the convolutional rendering for NeRF-GAN models. 

\label{sec:experiments}
\section{Conclusion}
We presented a method to distill a pretrained NeRF-GAN into a pose-conditioned convolutional generator. The proposed method enables considerably higher efficiency, which is crucial if 3D neural rendering is to become ubiquitous and deployed at scale. To do so, we proposed exploiting the intermediate latent space of the pretrained NeRF-GAN as a conditioning input of the convolutional generator. We additionally provided a training protocol to further improve the visual quality and 3D consistency of the images generated using our generator. Through our experiments, we showed that our method maintains good image quality and 3D consistency, significantly better than previous existing fully-convolutional methods and approaching that of the baseline NeRF-GAN with volumetric rendering. Finally, while our method takes steps toward achieving full correspondence between the two rendering methods, there remains a gap in terms of image semantics. Improving this aspect remains a subject for further research.
\label{sec:conclusion}

{\small
\bibliographystyle{ieee_fullname}
\bibliography{egbib}
}

\clearpage

\appendix

\section*{Supplementary Material}

In what follows, we first provide an evaluation of the correspondence between the generated images using volumetric rendering and our convolutional generator in section~\ref{correspondence}. Moreover, in section~\ref{efficiency}, we extend the efficiency analysis of our method to the setup of the Shapenet Cars dataset. In section~\ref{detail}, we discuss some of the implementation details. Then, we offer a discussion on the limitations of the proposed method in Sec.~\ref{limit}. Finally, we provide more visual examples for our method in Sec.~\ref{visual}.

\section{Evaluation of Correspondence}\label{correspondence}
To provide a baseline for future research, we measure the correspondence between generated images using our method and the corresponding images rendered from EG3D using Peak Signal to Noise Ratio (PSNR), the result of which is provided in Table \ref{tab_psnr}.

\section{Analysis of Efficiency on Shapenet Cars}\label{efficiency}
In Fig.~\ref{fig_supp_compute_cars}, we provide a comparison of the computational efficiency of inference using our method and the baseline NeRF-GAN (EG3D) on Shapenet Cars setup. As mentioned in section 4.3 of the main paper, EG3D performs volumetric rendering at the resolution of 64 and super-resolves the output to the resolution of 128. As it can be seen in Fig.~\ref{fig_supp_compute_cars}, Our method with StyleGAN2 as its backbone is more efficient than EG3D in terms of both memory consumption and speed. However, the StyleGAN3 backbone is only more efficient in terms of memory consumption, but it is slower than the baseline. The reason for this is that StyleGAN3 is more computationally demanding than the tri-plane generator in EG3D, which is based on StyleGAN2. In the setup of Shapenet Cars, the computational complexity of the StyleGAN3 convolutional generator outweighs that of the lower-resolution volumetric rendering in EG3D, resulting in a slower inference (Note that the memory consumption is still lower for the StyleGAN3 convolutional generator). 

\begin{figure}[t]
\begin{center}
\includegraphics[width=\linewidth]{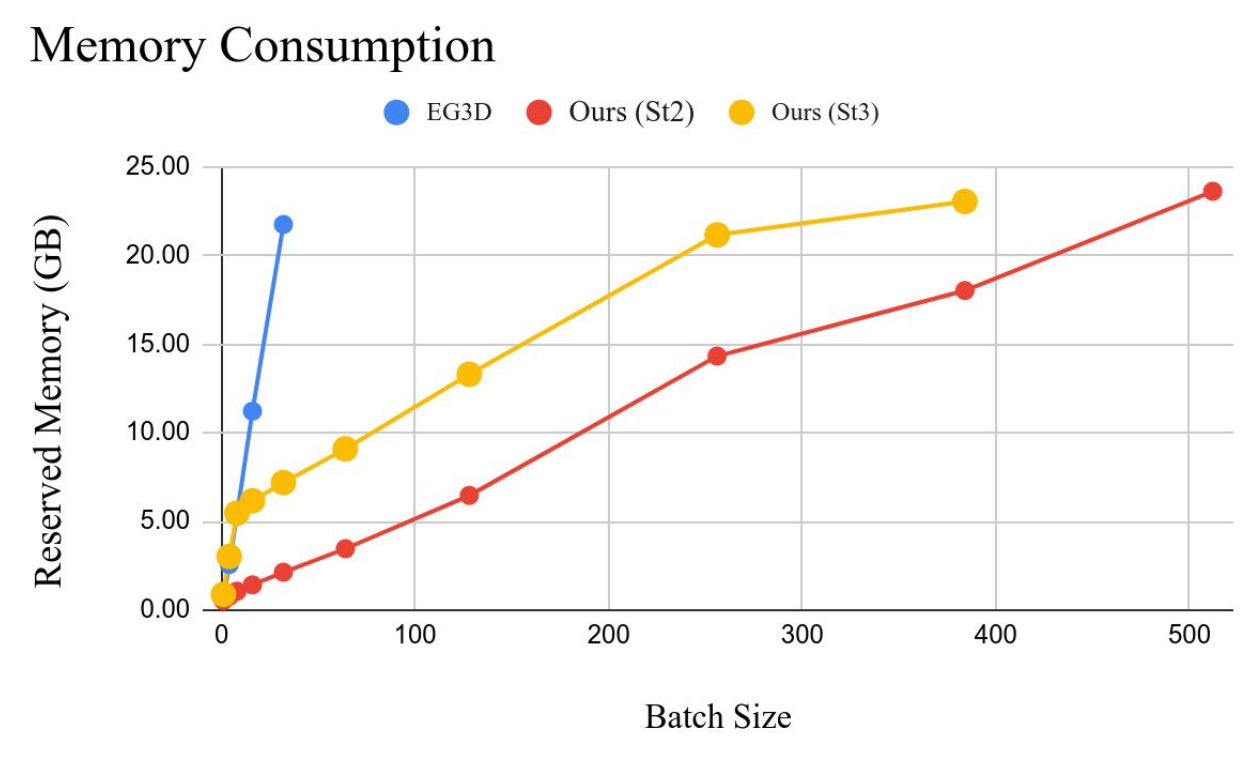}
\includegraphics[width=\linewidth]{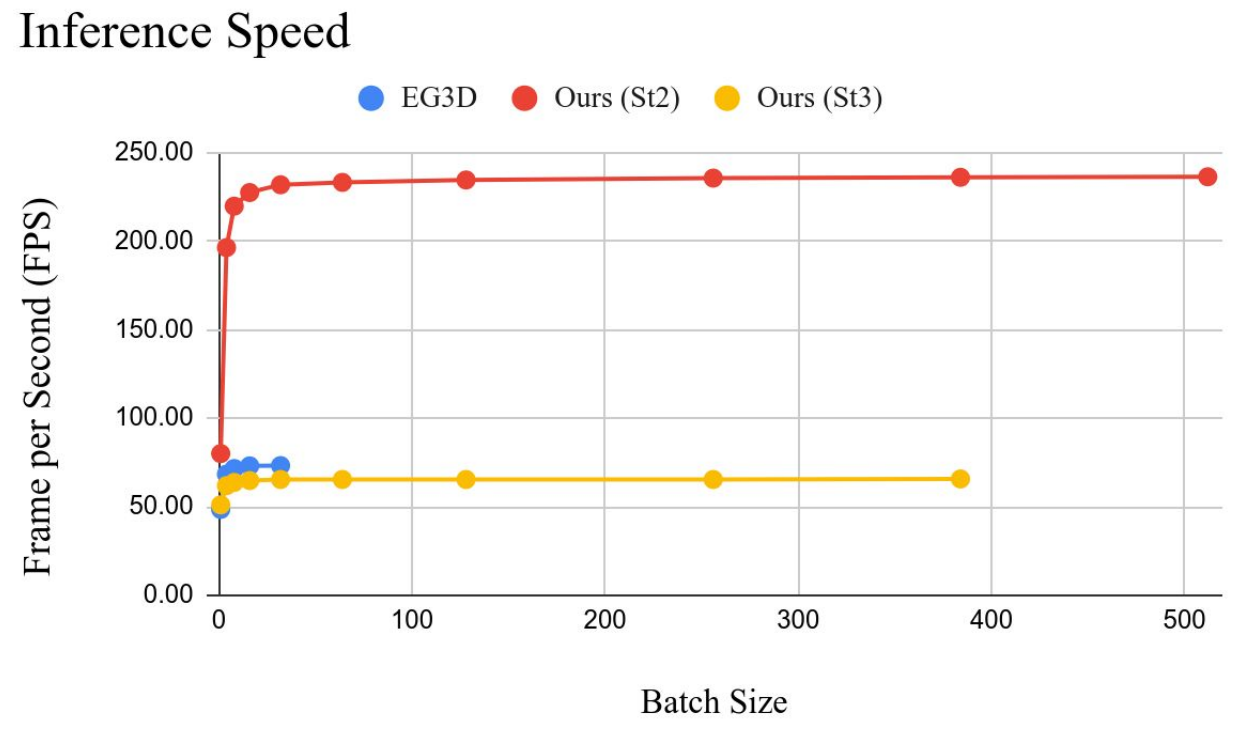}
\end{center}
\caption{Comparison of the inference memory consumption and speed (on a fixed GPU budget) for our method and EG3D on Shapenet Cars.}
\label{fig_supp_compute_cars}
\end{figure}

\begin{table}[b]
\caption{PSNR values ($\uparrow$) for measuring the correspondence between the images of EG3D and our convolutional generator on 3 datasets. The values are calculated for 1k images. ST2 and ST3 correspond to the two architectures StyleGAN2 and StyleGAN3 used as the backbone of our method}%
\label{tab_psnr}%
\centering
\resizebox{0.5\linewidth}{!}{%
\begin{tabular}{lll}
\multicolumn{1}{l}{Dataset} & \multicolumn{1}{l}{ST2} & \multicolumn{1}{l}{ST3}  \\
\toprule
\multicolumn{1}{l}{FFHQ} & \multicolumn{1}{l}{28.73} & \multicolumn{1}{l}{28.65}  \\
\multicolumn{1}{l}{AFHQ Cats} & \multicolumn{1}{l}{28.46} & \multicolumn{1}{l}{28.10}   \\
\multicolumn{1}{l}{Shapenet Cars} & \multicolumn{1}{l}{36.60} & \multicolumn{1}{l}{36.57} \\
\bottomrule
\end{tabular}}
\end{table}

\section{Implementation Details}\label{detail}

\subsection{SURF Baseline}\label{base}
In the main paper, we provided a baseline, which we referred to as ``SURF'', inspired by the method proposed in the SURF-GAN paper~\cite{kwak2022injecting} for pose/content disentanglement of a pre-trained 2D StyleGAN. As the code and the pre-trained models for such disentanglement are not made publicly available, we provided our best effort in implementing the SURF baseline inspired by their approach. 

The proposed method in~\cite{kwak2022injecting} consists of two main components: 1. a NeRF-GAN model called SURF-GAN for portrait images 2. a training recipe to add pose conditioning to a pre-trained StyleGAN2 using multi-view images generated from the 3D NeRF-GAN. Specifically, the part relevant as a baseline for our proposed method is the second component, which we tried to re-implement it for comparison with our method. To do so, for a fair comparison with our method, we use EG3D pre-trained on FFHQ for generating multi-view image triplets from 3 different viewpoints (source, canonical, and target views). Following ~\cite{kwak2022injecting}, we use pSp~\cite{richardson2021encoding}, a pretrained inversion encoder to invert the generated multi-view images in the style space of the 2D generator. Using the multi-view images and their corresponding style codes obtained using pSp, we train two mapping networks to: 
\begin{enumerate}[noitemsep]
    \item map any arbitrary style code to the style code of the canonical viewpoint, and
    \item map the canonical style code to the style code of the target viewpoint.
\end{enumerate}
  For the mapping networks, we use MLP networks with the same architecture as StyleGAN2's mapping network. For the training objective, we include reconstruction losses both on the images (MSE and Perceptual) and latent codes (MSE), as used in ~\cite{kwak2022injecting}. This architecture and training regime is inspired by the one in~\cite{kwak2022injecting}, but not exactly the same. We have adapted the method to the 3D generator network we are using (EG3D). Moreover, in the original method, the mapping from the canonical latent code to the target one is formulated as the linear combination of a set of learnable pose vectors. In our implementation, we use a more general mapping by using an MLP network instead. Nevertheless, the implemented baseline provides a comparison with a proven alternative strategy to enable pose conditioning of a StyleGAN-like convolutional architecture.

\subsection{Inversion}\label{inv}
In Sec. 4.8 of the main paper, we provided examples of inversion in our generator using Pivotal Tuning Inversion (PTI)~\cite{roich2022pivotal}. Following EG3D, inversion is performed given a target image and its corresponding camera parameters. As for the implementation, we use the adaptation to EG3D provided at~\cite{inversion}. We use 500 steps for optimizing the latent code and 350 additional steps for fine-tuning the generator according to the PTI method.

\subsection{3D Consistency Metrics}\label{3dmetric}
To calculate the 3D Landmark Consistency and Identity Preservation (ID) in section 4.5.3 of the main paper, we vary the yaw from -40 to +40 and the pitch from -30 to +30, following the evaluation setup of ~\cite{kwak2022injecting}. As an additional visualization, Fig.~\ref{fig_id} shows the comparison of Identitiy Preservation for each angle individually.

\begin{figure}
\begin{center}
\includegraphics[width=\linewidth]{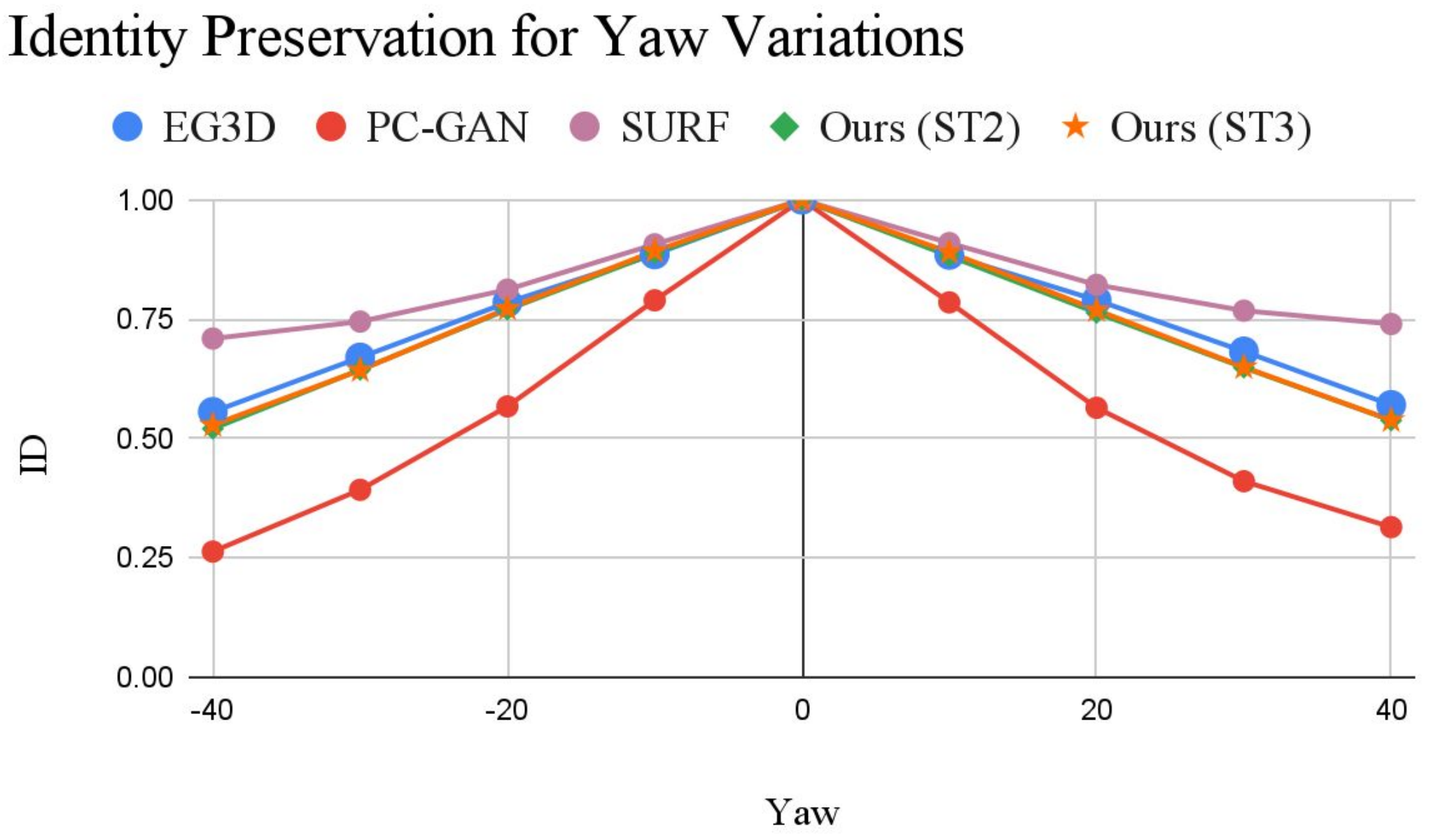}
\includegraphics[width=\linewidth]{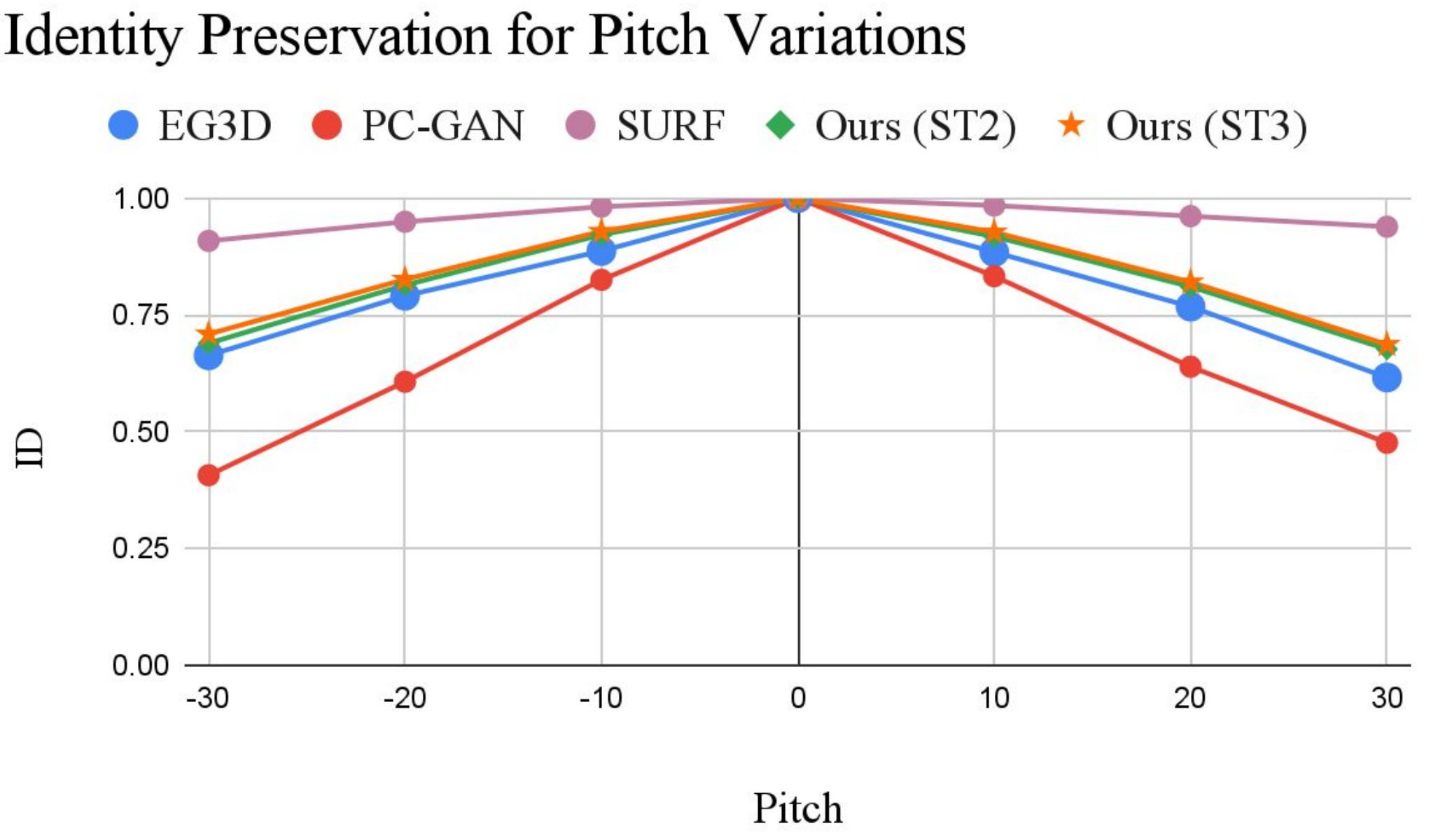}
\end{center}
\caption{Comparison of identity preservation across different camera poses. Our method performs much better than the pose-conditioned GAN baseline (PC-GAN), approaching the consistency of volumetric rendering (EG3D). The higher consistency achieved by SURF is due to its limited pose variations.}
\label{fig_id}
\end{figure}

 \section{Limitations}\label{limit}
In our approach, the outputs of the volumetric rendering branch are used to supervise our convolutional renderer. Thus, the visual quality and 3D consistency of our approach is largely bound by the quality and consistency of the pretrained NeRF-GAN. 
However, our formulation is largely agnostic to the volumetric generator used. Therefore, improvements in volumetric rendering in the context of GANs will also transfer to the generated quality of our approach.

As an example, a close look at the videos provided in the supplementary video reveals small changes in the degree of the smile in different viewpoints of some of the generated faces. Such a phenomenon, which is due to the dataset bias, also exists in EG3D, as analyzed in Sec 1.1 of the supplementary material of EG3D paper~\cite{chan2022efficient} and visible in our results as well.

\section{Visual Results}\label{visual}
In Fig.~\ref{fig_supp_ffhq}, we provide more visual examples generated using our method from FFHQ dataset. Fig.~\ref{fig_supp_cats} shows examples of generated images from AFHQ Cats dataset using the proposed convolutional generator. Finally, in Fig.~\ref{fig_supp_cars}, we provide random samples from Shapenet Cars dataset generated using the volumetric rendering (EG3D) and our method. For a video comparison, please refer to the supplementary video ``\textcolor{red}{supp.mp4}''.

\begin{figure*}[t]
\begin{center}
\includegraphics[width=0.95\linewidth]{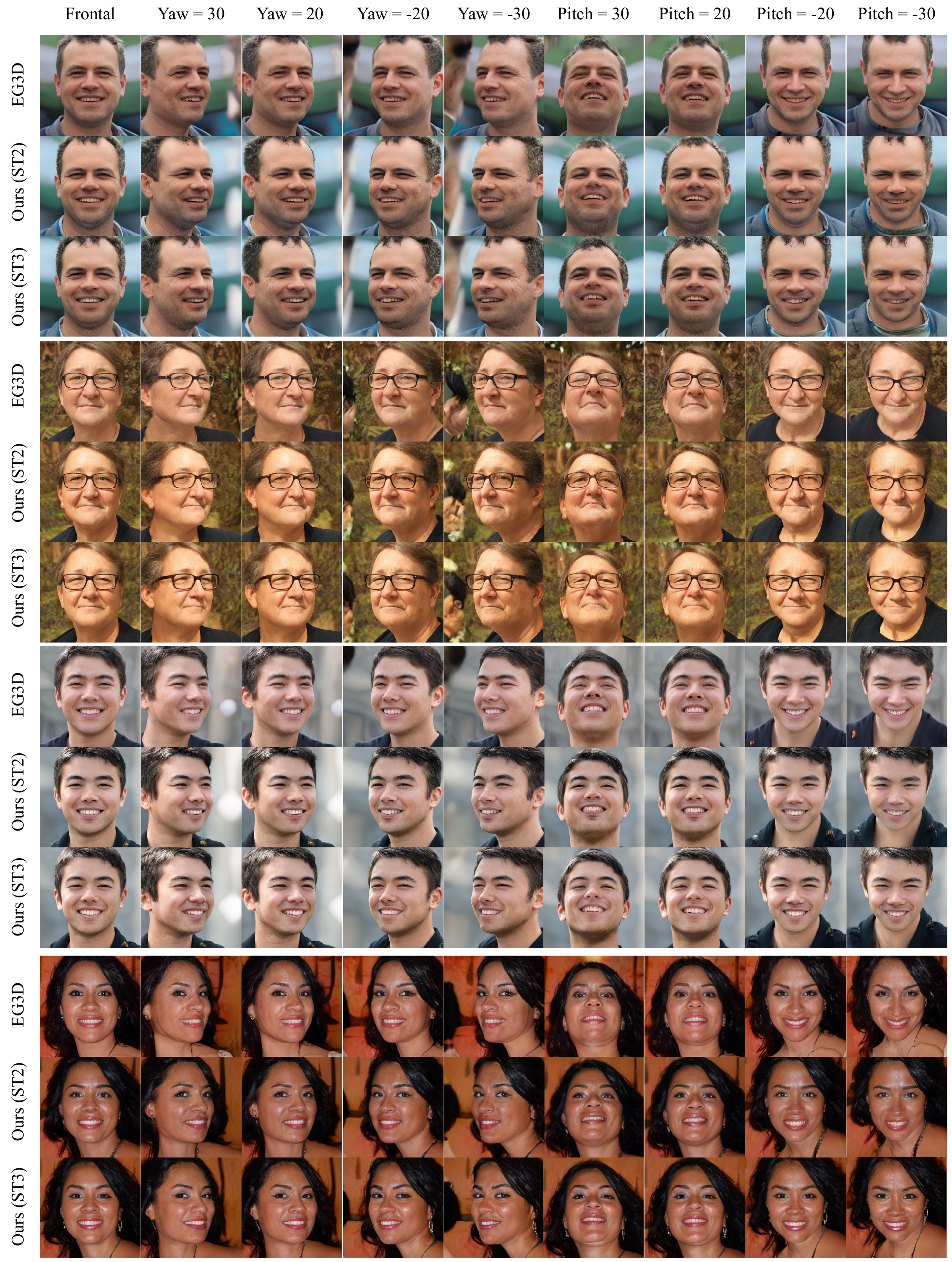}
\end{center}
\vspace{-5mm}
\caption{Visual examples of pose control in our convolutional generator and their comparison to those of EG3D on FFHQ dataset.}
\label{fig_supp_ffhq}
\end{figure*}

\begin{figure*}[t]
\begin{center}
\includegraphics[width=0.95\linewidth]{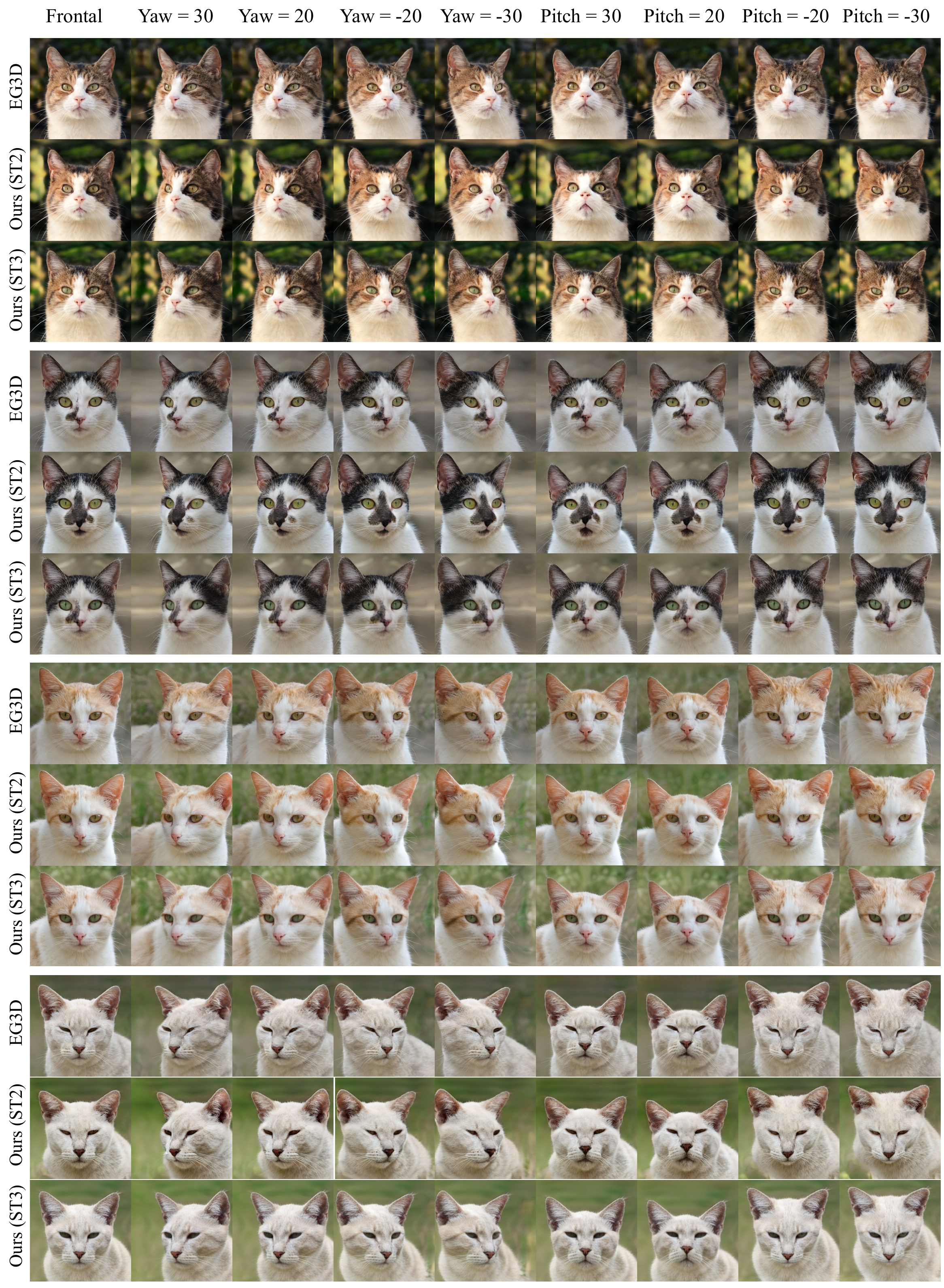}
\end{center}
\vspace{-5mm}
\caption{Visual examples of pose control in our convolutional generator and their comparison to those of EG3D on AFHQ Cats dataset.}
\label{fig_supp_cats}
\end{figure*}

\begin{figure*}[t]
\begin{center}
\includegraphics[width=0.8\linewidth]{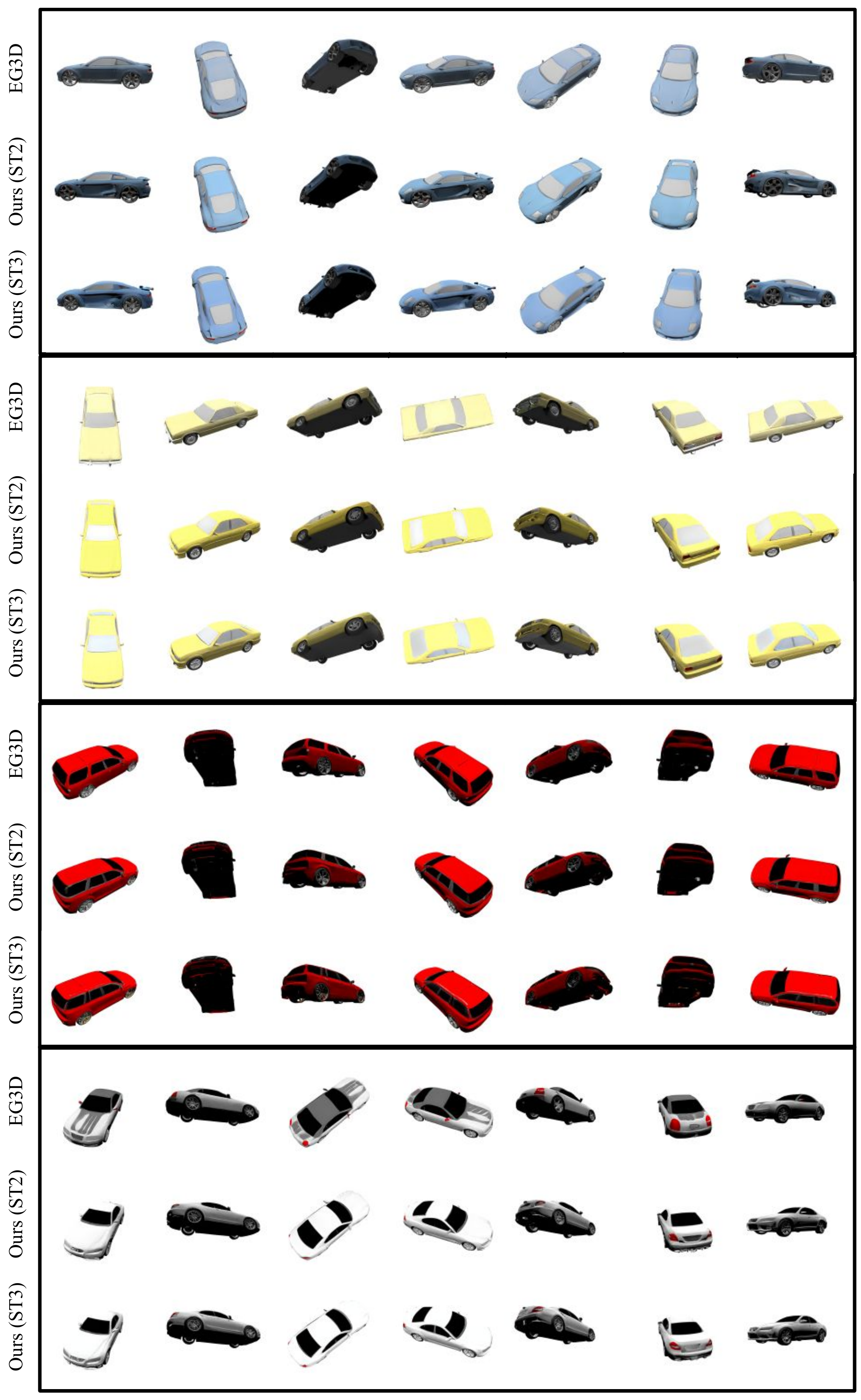}
\end{center}
\vspace{-5mm}
\caption{Visual examples of pose control in our convolutional generator and their comparison to those of EG3D on Shapenet Cars dataset.}
\label{fig_supp_cars}
\end{figure*}


\end{document}